\documentclass[letterpaper, 10 pt, conference]{ieeeconf}

\IEEEoverridecommandlockouts
\usepackage[T1]{fontenc}
\usepackage{times}
\usepackage{amsmath}
\usepackage{amssymb}
\usepackage{graphicx}
\usepackage[ruled, linesnumbered]{algorithm2e}
\usepackage{booktabs}
\usepackage{float}
\usepackage{bigstrut}
\usepackage{multirow}
\usepackage{balance}
\usepackage[hidelinks=true,bookmarks=false,colorlinks=True]{hyperref}
\usepackage{caption}
\usepackage{lineno}
\usepackage{amsopn}
\usepackage{comment}
\usepackage{color}
\usepackage{cite}
\usepackage{algorithmic}
\usepackage{tabularx}
\usepackage{xcolor}

\SetArgSty{textnormal}

\makeatletter

\newcommand{\Rmnum}[1]{\expandafter\@slowromancap\romannumeral #1@}
\makeatother

\newcolumntype{L}[1]{>{\raggedright\arraybackslash}p{#1}}
\newcolumntype{C}[1]{>{\centering\arraybackslash}p{#1}}
\newcolumntype{R}[1]{>{\raggedleft\arraybackslash}p{#1}}

\usepackage{pifont}
\newcommand{\cmark}{\ding{51}}%

\title{\LARGE \bf S2P2: Self-Supervised Goal-Directed Path Planning Using RGB-D Data for Robotic Wheelchairs}

\author{Hengli Wang, Yuxiang Sun, Rui Fan, and Ming Liu, \IEEEmembership{Senior Member, IEEE}
\thanks{\textit{(Corresponding author: Ming Liu.)}}
\thanks{Hengli Wang and Ming Liu are with the Department of Electronic and Computer Engineering, The Hong Kong University of Science and Technology, Clear Water Bay, Kowloon, Hong Kong SAR, China (email: hwangdf@connect.ust.hk; eelium@ust.hk).}
\thanks{Yuxiang Sun is with the Department of Mechanical Engineering, The Hong Kong Polytechnic University, Hung Hom, Kowloon, Hong Kong (e-mail:
yx.sun@polyu.edu.hk, sun.yuxiang@outlook.com).}
\thanks{Rui Fan is with the Department of Computer Science and Engineering, and the Department of Ophthalmology, the University of California San Diego, La Jolla, CA 92093, United States (email: rui.fan@ieee.org).}
}

\begin{document}

\maketitle
\thispagestyle{empty}
\pagestyle{empty}

\begin{abstract}
Path planning is a fundamental capability for autonomous navigation of robotic wheelchairs. With the impressive development of deep-learning technologies, imitation learning-based path planning approaches have achieved effective results in recent years. However, the disadvantages of these approaches are twofold: 1) they may need extensive time and labor to record expert demonstrations as training data; and 2) existing approaches could only receive high-level commands, such as turning left/right. These commands could be less sufficient for the navigation of mobile robots (\textit{e.g.,} robotic wheelchairs), which usually require exact poses of goals. We contribute a solution to this problem by proposing S2P2, a self-supervised goal-directed path planning approach. Specifically, we develop a pipeline to automatically generate planned path labels given as input RGB-D images and poses of goals. Then, we present a best-fit regression plane loss to train our data-driven path planning model based on the generated labels. Our S2P2 does not need pre-built maps, but it can be integrated into existing map-based navigation systems through our framework. Experimental results show that our S2P2 outperforms traditional path planning algorithms, and increases the robustness of existing map-based navigation systems. Our project page is available at \url{https://sites.google.com/view/s2p2}.
\end{abstract}

\section{Introduction}
\label{sec.introduction}
Robotic wheelchairs play a significant role in improving the mobility of disabled and elderly people. Autonomous navigation is an essential capability for robotic wheelchairs \cite{wang2021dynamic}. To achieve it, traditional approaches typically consist of three modules: perception, path planning and control \cite{elfes1989using,fan2020sne,liu2021the}. Although decomposing the whole system into individual modules allows each module to be developed independently, it may suffer from the accumulation of the uncertainties of each module. For example, inaccurate perception results may lead to unsafe paths or detours.

To address this problem, Levine \textit{et al.} \cite{levine2016end} showed that training the perception and control modules jointly could allow individual modules to cooperatively improve the overall performance. Recently, many researchers have resorted to the end-to-end control paradigm \cite{hecker2018end,bojarski2017explaining,pfeiffer2017perception}, which integrates perception, path planning and control into one module. However, Xu \textit{et al.} \cite{xu2017end} argued that the learned control policy would be limited to specific actuation setups or the simulation environments in which the training was performed.

\begin{figure}[t]
   \centering
   \includegraphics[width=0.8\linewidth]{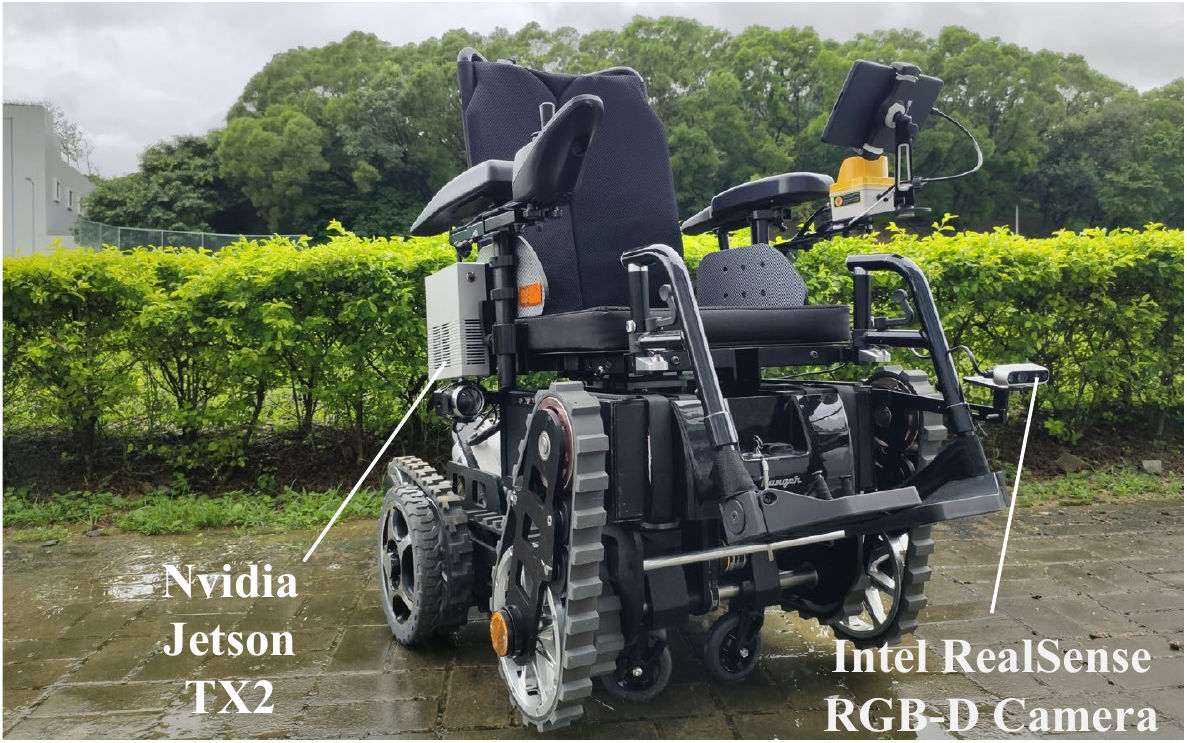}
   \caption{The robotic wheelchair used in this work. It is equipped with an Intel RealSense RGB-D camera to collect data and an NVIDIA Jetson TX2 to run our S2P2 model.}
   \label{fig.wheelchair}
\end{figure}

Different from the above-mentioned approaches, many end-to-end path planning approaches based on imitation learning have been developed, which take as input raw sensor data and output planned paths instead of control signals to make the model more generic \cite{thesis,bansal2018chauffeurnet,cai2020vtgnet,sun2020see,wang2021learning}. However, these approaches generally have two disadvantages: 1) experts are often required to drive robots many times in various scenes, so extensive time and labor may be needed to record expert demonstrations as training data; and 2) existing imitation learning-based approaches could only receive high-level commands at runtime, such as turning left/right or going straight \cite{cai2020vtgnet}. These commands could be less sufficient for the navigation of mobile robots (\textit{e.g.,} our robotic wheelchair shown in Fig.~\ref{fig.wheelchair}), which usually require exact poses of goals. For example, given a right-turning command,  a robotic wheelchair in a large free space could have many options to go to the right side. The simple high-level commands may not accurately direct the robot to the goal.

\begin{figure}[t]
   \centering
   \includegraphics[width=0.99\linewidth]{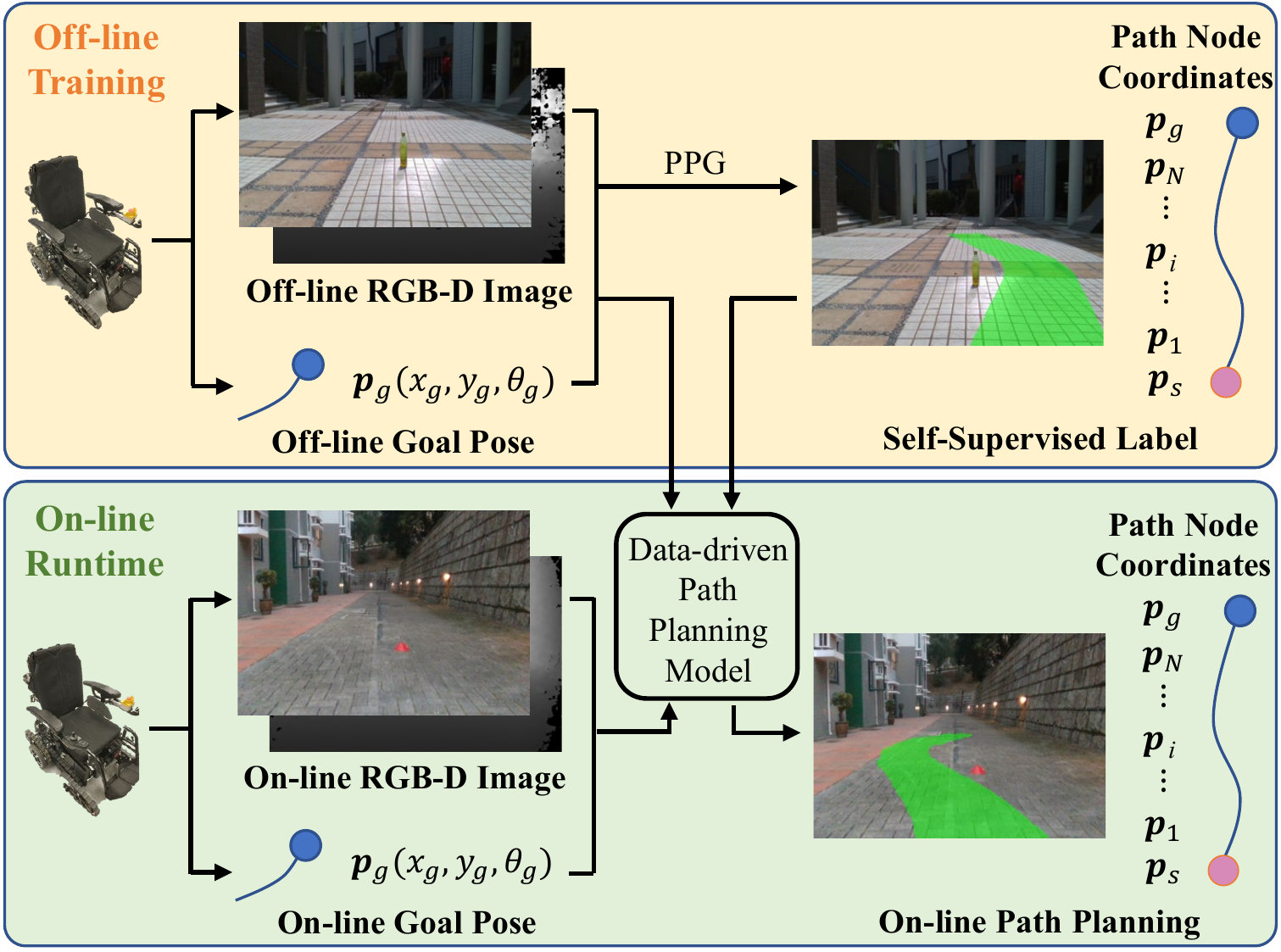}
   \caption{An overview of our S2P2. We first use our proposed PPG to generate self-supervised labels (top), which are then used to train the data-driven path planning model based on our best-fit regression plane loss. At runtime, a robotic wheelchair equipped with an RGB-D camera can perform the on-line path planning (bottom).}
   \label{fig.framework}
\end{figure}

To tackle the above issues, we present S2P2, a \underline{S}elf-\underline{S}upervised goal-directed \underline{P}ath \underline{P}lanning approach for robotic wheelchairs, using an Intel RealSense RGB-D camera. The main difference between our S2P2 and other networks is that our S2P2 does not need manual demonstrations and can directly take as input poses of goals instead of high-level commands. Specifically, we adopt the end-to-end path planning paradigm, which directly takes as input RGB-D images as well as poses of goals, and outputs planned paths.

Fig.~\ref{fig.framework} illustrates the overview of our S2P2. We first develop a pipeline named the planned path generator (PPG) to automatically generate planned path labels given as input RGB-D images \cite{sun2017improving,sun2018motion,sun2019active} and poses of goals. Then, we present a best-fit regression plane loss to train our proposed data-driven path planning model based on the generated labels. We also propose a framework that allows our mapless S2P2 to be integrated into existing map-based navigation systems. Experimental results show that our S2P2 outperforms traditional path planning algorithms, and increases the robustness of existing map-based navigation systems. The contributions of this paper are summarized as follows:

\begin{enumerate}
\item We develop S2P2, which contains an automatic labeling pipeline named PPG, a novel best-fit regression plane loss and a data-driven path planning model.
\item We propose a framework allowing our mapless S2P2 to be integrated with any map-based navigation system.
\item Experimental results demonstrate the superiority of both our S2P2 and the S2P2-integrated navigation system.
\end{enumerate}

\section{Related Work}

\subsection{Traditional Path Planning Approaches}
Traditional path planning algorithms could be generally divided into two categories, complete algorithms and sampling-based algorithms. Complete algorithms, such as A* \cite{Astar} and JPS \cite{JPS}, could always find a solution if one exists, but they are computationally intensive. Sampling-based algorithms, such as PRM \cite{PRM} and RRT* \cite{RRTstar}, trade off between the quality of planned paths and efficiency. These algorithms may fail due to the uncertainties from the perception module.

To address this problem, many researchers have proposed to plan under obstacle uncertainties \cite{kuwata2012risk,ono2015chance,jasour2019risk}. Kuwata \textit{et al.} \cite{kuwata2012risk} formulated this problem as a chance-constrained dynamic programming problem, and solved it by performing cost analysis. Recently, Jasour \textit{et al.} \cite{jasour2019risk} proposed a novel risk-contour map and employed this map to obtain safe paths for robots with guaranteed bounded risks.

\subsection{End-to-end Path Planning Approaches}
There are many studies that could output the coordinates of the planned paths via imitation learning \cite{thesis,bansal2018chauffeurnet,cai2020vtgnet,sun2020see,wang2021learning}. Bergqvist \textit{et al.} \cite{thesis} compared different combinations of convolutional neural networks (CNNs) and long short-term memory (LSTM) networks, and concluded that the path planned by LSTM or CNN-LSTM is smooth and feasible in many situations. Cai \textit{et al.} \cite{cai2020vtgnet} proposed a vision-based model, which receives camera images to plan a collision-free trajectory in the future. These imitation learning-based approaches, however, have two disadvantages as mentioned in Section~\ref{sec.introduction}, which greatly limit their applications.

\subsection{End-to-end Control Approaches}
ALVINN \cite{pomerleau1989alvinn} was the first attempt to implement end-to-end control for mobile robots by using a shallow neural network. Inspired by ALVINN, Bojarski \textit{et al.} \cite{bojarski2017explaining} proposed PilotNet, which utilizes CNNs to map front-view images directly to steering commands. There also exist studies that utilize other sensors (\textit{e.g.,} LiDARs) besides cameras \cite{pfeiffer2017perception,hecker2018end}. However, these end-to-end control approaches exhibit low generalization capabilities as already mentioned \cite{xu2017end}.

\begin{figure*}[t]
   \centering
   \includegraphics[width=0.99\textwidth]{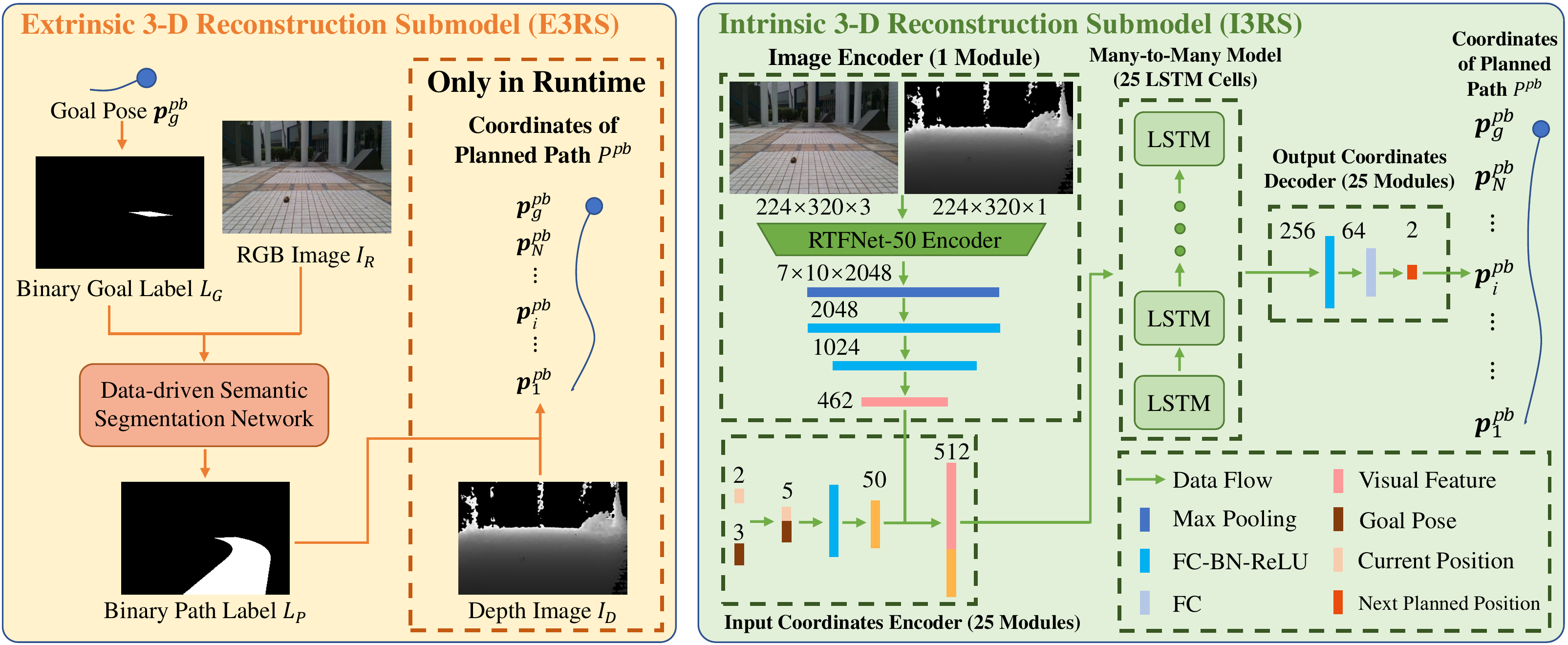}
   \caption{An overview of our data-driven path planning model, which consists of two different submodels. E3RS (left) transforms the original path planning problem to a semantic segmentation problem. At runtime, the coordinates of the planned path $P^{pb}$ can be computed by extrinsically reconstructing the binary path label $L_{P}$. I3RS (right) utilizes a CNN-LSTM neural network to output $P^{pb}$ intrinsically. Each LSTM module takes as input the combined feature including the current position information, and outputs a feature that is then decoded to the next planned position.}
   \label{fig.DPPM}
\end{figure*}

\section{Methodology}

\subsection{Problem Formulation}
Let $(\cdot)^{w}$ denote the world frame, $(\cdot)^{b}$ denote the robot body frame and $(\cdot)^{c}$ denote the camera frame. $\mathbf{R}_{b}^{w}$ and $\mathbf{T}_{b}^{w}$ represent the rotation matrix and the translation matrix from the body frame to the world frame, respectively. Given an input registered front-view RGB image $I_{R}$ and depth image $I_{D}$, we can construct a 3-D point cloud of the front scene, which contains the configuration space of the robotic wheelchair. To simplify the analysis, we set the configuration space $C \subset \mathbb{R}^{2} \times \mathrm{SO}(2)$, where $\mathrm{SO}(2)$ denotes the 2-D rotation group. Since the $z$-axis value has no impact on the path planning of robotic wheelchairs, we consider a 2-D projected body frame denoted by $(\cdot)^{pb}$, which coincides with the $x-y$ plane of the body frame.

We consider the problem of generating a path given any arbitrary goal pose $\boldsymbol{p}_{g}^{pb} \in C$ without colliding with obstacles. Then, the generated path can be expressed as:
\begin{equation}
   P^{pb} \triangleq\left\{\boldsymbol{p}_{1}^{pb}, \ldots, \boldsymbol{p}_{i}^{pb}, \ldots, \boldsymbol{p}_{N}^{pb}, \boldsymbol{p}_{g}^{pb}\right\},
   \label{eq.path}
\end{equation}
\noindent where $\boldsymbol{p}_{i}^{pb} \in C$ denotes a node of the planned path. In this paper, we set $N=24$. Note that we take the orientation of the goal pose $\theta_{g}^{pb}$ into consideration because it could affect the generated paths. However, our S2P2 does not output the orientation of the generated path nodes because the subsequent control module only needs the position information.

\subsection{Planned Path Generator}
Our PPG is designed to automatically generate planned path labels $P^{pb}$ given as input RGB-D images and poses of goals. Given $I_{R}$ and $I_{D}$, we first use \cite{wang2019self} to generate corresponding semantic segmentation image $I_{S}$, which can provide pixel-level predictions of the drivable area and obstacles. For any pixel $\boldsymbol{q}$ in the image, we calculate its 3-D coordinate in the camera frame $\boldsymbol{q}^{c}$ by using $I_{D}$. Then, we can obtain the point cloud in the camera frame $PC^{c}$ by calculating the 3-D coordinates of all pixels. We further employ \cite{rusu2008towards} to filter out outliers in $PC^{c}$. Afterwards, we build an occupancy grid map named costmap $M^{pb}$ in the projected body frame. The initialization approach is described in Algorithm \ref{algo.costmap}, where $\boldsymbol{q}_{d}^{pb}$ and $\boldsymbol{q}_{o}^{pb}$ denote the point belonging to the drivable area and obstacles, respectively. For any point $\boldsymbol{q}^{c} \in PC^{c}$, we calculate its coordinate in the projected body frame $\boldsymbol{q}^{pb}$ via $\mathbf{R}_{c}^{b}$ and $\mathbf{T}_{c}^{b}$ (line 1). We then constrict the free area (line 7) and inflate the occupied area (line 8). Since the size of our robotic wheelchair is $1m \times 0.5m$, we accordingly set two radiuses both as $0.5m$. The size of each cell in $M^{pb}$ is $0.1m \times 0.1m$.

\begin{figure*}[t]
   \centering
   \includegraphics[width=0.99\linewidth]{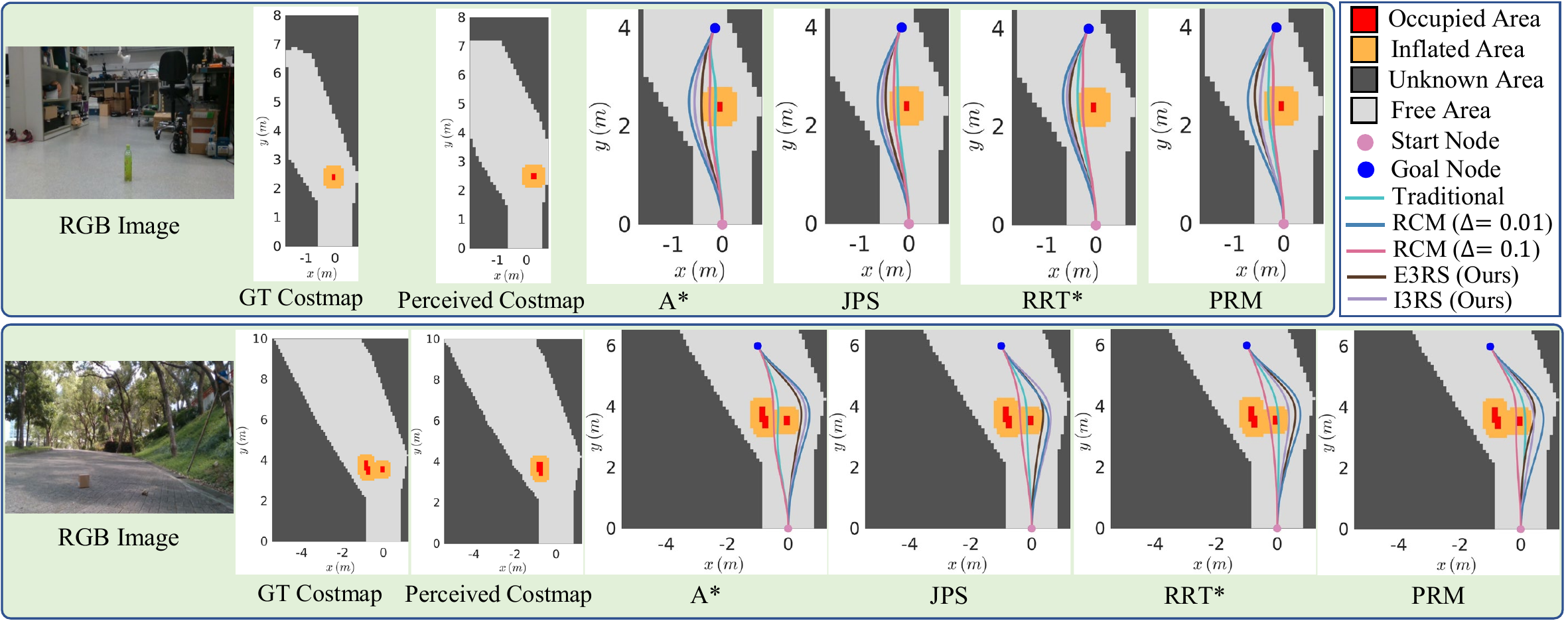}
   \caption{Two examples (indoor and outdoor) with inaccurate perception results for the comparison between traditional algorithms, RCM \cite{jasour2019risk} with two different risk bounds, and our proposed E3RS and I3RS. Perceived costmaps and ground-truth costmaps are constructed by the semantic predictions and ground-truth semantic segmentation labels, respectively. The paths planned by different approaches are all displayed on the ground-truth costmaps. Although the perception results are not entirely accurate, our proposed E3RS and I3RS can still present a better performance than the other approaches.}
   \label{fig.result}
\end{figure*}

Now, given a goal pose $\boldsymbol{p}_{g}^{pb}$, we can plan a path $P^{pb}$ in $M^{pb}$ by using traditional path planning algorithms. If $\boldsymbol{p}_{g}^{pb}$ lies outside the free area due to perception or localization errors, we will take the closest point to $\boldsymbol{p}_{g}^{pb}$ in the free area as the new goal and replan the path. After obtaining $P^{pb}$ from the above steps, we project the input goal pose and the planned path to the original image and obtain the binary path label $L_{P}$ and binary goal label $L_{G}$, as presented in Fig.~\ref{fig.DPPM}. By sampling goal poses randomly, our proposed PPG can generate a large number of planned path labels given as input RGB-D images, which saves much time and labor.

\begin{algorithm}[t]
   \KwIn{$PC^{c}$.}
   \KwOut{$M^{pb}$.}
   Compute $\boldsymbol{q}^{pb}$ for every point $\boldsymbol{q}^{c} \in PC^{c}$\\
   Initialize $M^{pb}$ as \textbf{unknown area}\\
   Compute region $R_{d}^{pb} \gets \textbf{findConvexHull}\left(\boldsymbol{q}_{d}^{pb}\right)$\\
   Set $M^{pb}\left(R_{d}^{pb}\right)$ as \textbf{free area}\\
   Compute region $R_{o}^{pb} \gets \textbf{findConvexHull}\left(\boldsymbol{q}_{o}^{pb}\right)$\\
   Set $M^{pb}\left(R_{o}^{pb}\right)$ as \textbf{occupied area}\\
   $\textbf{ConstrictFreeArea}\left(M^{pb}\right)$\\
   $\textbf{InflateOccupiedArea}\left(M^{pb}\right)$
   \caption{Costmap Initialization}
   \label{algo.costmap}
\end{algorithm}

\subsection{Data-driven Path Planning Model}
Although our PPG can generate the coordinates of the planned paths, the generated paths often present detours or unsafe routes due to perception errors. Therefore, we propose a data-driven path planning model to provide better planned paths given as input RGB-D images and poses of goals. The overview of our model is illustrated in Fig. \ref{fig.DPPM}.

Since the nodes of the planned paths are in sequential order, our model should be able to model this relationship. To this end, we propose two different submodels. The extrinsic 3-D reconstruction submodel (E3RS) analogizes this sequential relationship to the spatial continuity of an image and transforms this problem to a semantic segmentation problem, while the intrinsic 3-D reconstruction submodel (I3RS) utilizes LSTM to model this sequential relationship.

\subsubsection{Extrinsic 3-D Reconstruction Submodel (E3RS)}
We first project the input goal pose $\boldsymbol{p}_{g}^{pb}$ to the image frame and obtain the binary goal label $L_{G}$. Then, we train an existing data-driven semantic segmentation network, RTFNet-50 \cite{rtfnet}, that takes as input the binary goal label $L_{G}$ and RGB image $I_{R}$, and outputs binary path label $L_{P}$. At runtime, we use $I_D$ to compute $P^{pb}$ for the subsequent control module.

\subsubsection{Intrinsic 3-D Reconstruction Submodel (I3RS)}

\begin{table*}[t]
   \centering
   \caption{Performance comparison between traditional algorithms, RCM \cite{jasour2019risk} with two different risk bounds, and our E3RS and I3RS. $\uparrow$ means higher numbers are better, and $\downarrow$ means lower numbers are better. The best results are bolded.}
   \begin{tabular}{C{0.9cm}C{0.4cm}C{0.4cm}C{0.4cm}C{0.4cm}C{0.4cm}C{0.4cm}C{0.4cm}C{0.4cm}C{0.4cm}C{0.4cm}C{0.4cm}C{0.4cm}C{0.4cm}C{0.4cm}C{0.4cm}C{0.4cm}C{0.4cm}C{0.4cm}C{0.4cm}C{0.4cm}}
   \toprule
   \multirow{2}{*}{\begin{tabular}[c]{@{}c@{}}Evaluation\protect\\Metrics\end{tabular}} & \multicolumn{4}{c}{Traditional (PPG)} & \multicolumn{4}{c}{E3RS (\textbf{Ours})} & \multicolumn{4}{c}{I3RS (\textbf{Ours})} & \multicolumn{4}{c}{{RCM ($\Delta=0.01$)}} & \multicolumn{4}{c}{{RCM ($\Delta=0.1$)}} \\ \cmidrule(l){2-5} \cmidrule(l){6-9} \cmidrule(l){10-13} \cmidrule(l){14-17} \cmidrule(l){18-21}
    & A* & JPS & RRT* & PRM & A* & JPS & RRT* & PRM & A* & JPS & RRT* & PRM & {A*} & {JPS} & {RRT*} & {PRM} & {A*} & {JPS} & {RRT*} & {PRM} \\ \midrule
   SR $\uparrow$ & 70.7$\%$ & 65.6$\%$ & 70.4$\%$ & 68.6$\%$ & 89.2$\%$ & 86.9$\%$ & \textbf{91.7$\%$} & 88.1$\%$ & 90.2$\%$ & 87.9$\%$ & 89.6$\%$ & 86.4$\%$ & {87.8$\%$} & {85.7$\%$} & {86.5$\%$} & {83.6$\%$} & {69.6$\%$} & {67.3$\%$} & {70.2$\%$} & {65.9$\%$} \\
   TC $\downarrow$ & 0.132 & 0.151 & 0.127 & 0.187 & 0.115 & 0.116 & 0.089 & 0.096 & 0.091 & 0.121 & \textbf{0.082} & 0.125 & {0.159} & {0.167} & {0.131} & {0.192} & {0.137} & {0.152} & {0.125} & {0.179} \\
   \bottomrule
   \end{tabular}
   \label{tab.result}
\end{table*}

We construct a CNN-LSTM network to model the sequential relationship between each node of the planned path. We first use the encoder of RTFNet-50 \cite{rtfnet} to extract the visual features of a given RGB-D image. To balance the dimensional difference between the coordinate vector and visual features, we encode the input coordinate vector to a feature that has a higher dimension, which is then concatenated with the visual features and fed to the LSTM cell. Each LSTM cell takes the combined feature as input, and outputs a feature that is then decoded to the next planned position. The current position feature fed to one LSTM cell comes from the next planned position feature given by the previous LSTM cell, and the first LSTM cell takes $(0, 0)$ as the current position. Since $P^{pb}$ contains 25 nodes including the input goal node, there are a total of 25 LSTM cells that constitute our many-to-many LSTM model.

\begin{algorithm}[t]
    \KwIn{$P^{w}, r$.}
    \KwOut{$G^{w}$.}
    $G^{w} \gets \emptyset$\\
    $cur \gets \boldsymbol{p}_{s}^{w}$\\
    \For{$i = 1 \to M$}
    {
       \If{$\left(\textbf{Visible}\left(\boldsymbol{p}_{i}^{w}, cur\right) \wedge \neg \textbf{Visible}\left(\boldsymbol{p}_{i+1}^{w}, cur\right)\right) \vee$ \newline
       $\left(\textbf{Dist}\left(\boldsymbol{p}_{i+1}^w, cur\right)>r\right)$}
       {
          $G^{w}.\textbf{insert}\left(\boldsymbol{p}_{i}^{w}\right)$\\
          $cur \gets \boldsymbol{p}_{i}^{w}$
       }
    }
    $G^{w}.\textbf{insert}\left(\boldsymbol{p}_{g}^{w}\right)$
    \caption{Intermediate Goal Pose Generator}
    \label{algo.IGPG}
\end{algorithm}

\subsubsection{Training Loss for Our E3RS and I3RS}
The training loss for our E3RS $\mathcal{L}_{E}$ consists of two terms:
\begin{equation}
    \mathcal{L}_{E} = \mathcal{L}_{EP} + \lambda_{ER} \mathcal{L}_{ER},
\end{equation}
where $\mathcal{L}_{EP}$ and $\mathcal{L}_{ER}$ denote the PPG guiding loss and the best-fit regression plane loss, respectively. $\mathcal{L}_{EP}$ takes the binary path label generated by PPG $\widehat{L}_{P}$ as the supervision, and is defined as the cross entropy between $\widehat{L}_{P}$ and the E3RS prediction ${L}_{P}$. Moreover, since we regard the desired path as a plane, we design $\mathcal{L}_{ER}$ to penalize off-plane pixels with high probability in ${L}_{P}$. Referring to the generalized v-disparity analysis discussed in \cite{fan2019real,fan2019pothole}, the inverse depth (or disparity) pixels $1/I_D(\boldsymbol{q})$ are typically projected as a non-linear pattern $f(\boldsymbol{a},\boldsymbol{q},\phi)=a_0+a_1(-u\sin\phi+v\cos\phi)$ in the $v$ (vertical) direction \cite{fan2021learning,fan2021rethinking}, where $\boldsymbol{q}=[u,v]^T$ denotes the pixel and $\phi$ denotes the RGB-D camera roll angle. $\boldsymbol{a}=[a_0,a_1]^T$ and $\phi$ can be yielded by finding the best-fit regression plane, which corresponds to the minimum of the mean of squared residuals between the non-linear pattern and the pixels $\boldsymbol{q}$ with high probability in $L_P$ \cite{fan2019road,fan2020we}:
\begin{equation}
\mathcal{L}_{ER}=\frac{1}{N_p}\sum_{i=1}^{N_p}(1/I_D(\boldsymbol{q}_i)-f(\hat{\boldsymbol{a}},\boldsymbol{q}_i,\hat{\phi}))^2,
\label{eq.loss_er}
\end{equation}
where $N_p$ represents the number of pixels with a probability larger than 0.5 in $L_P$; and $\hat{\boldsymbol{a}}$ and $\hat{\phi}$ separately denote the optimum $\boldsymbol{a}$ and ${\phi}$. Their closed-form solutions are provided in \cite{fan2019road}. The off-plane pixels with high probability in $L_P$ can produce a relatively high $\mathcal{L}_{ER}$, and vice versa.

Similarly, the training loss for our I3RS $\mathcal{L}_{I}$ also consists of two terms:
\begin{equation}
    \mathcal{L}_{I} = \mathcal{L}_{IP} + \lambda_{IR} \mathcal{L}_{IR},
\end{equation}
where $\mathcal{L}_{IP}$ and $\mathcal{L}_{IR}$ denote the PPG guiding loss and the best-fit regression plane loss, respectively. $\mathcal{L}_{IP}$ takes the planned path generated by PPG $\widehat{P}^{pb}$ as the supervision:
\begin{equation}
    \mathcal{L}_{IP} = \frac{1}{25} \sum_{i=1}^{25} \left\| \widehat{\boldsymbol{p}}_{i}^{pb} - \boldsymbol{p}_{i}^{pb} \right\|_2,
\end{equation}
where $\widehat{\boldsymbol{p}}_{i}^{pb}$ and $\boldsymbol{p}_{i}^{pb}$ denote the planned path node via PPG and I3RS, respectively; and $||\cdot||$ denotes the $L2$-Norm. Moreover, we use the I3RS prediction $P^{pb}$ and $I_D$ to compute $\mathcal{L}_{IR}$ based on \eqref{eq.loss_er}. $\mathcal{L}_{IR}$ is also employed to penalize the on-path but off-plane points.

\subsection{The Proposed Framework for Integrating S2P2 into Existing Map-based Navigation Systems}
\label{sec:framework}
For long-range autonomous navigation tasks, the input goals are often outside the field of view (FOV) of the front-view camera. To address this problem, we propose a framework allowing our mapless S2P2 to be integrated into existing map-based navigation systems.

Given a goal pose, we can first plan a global collision-free path in the world frame $P^{w}=\left\{\boldsymbol{p}_{s}^{w}, \boldsymbol{p}_{1}^{w}, \ldots, \boldsymbol{p}_{M}^{w}, \boldsymbol{p}_{g}^{w}\right\}$ using global path planners (\textit{e.g.,} PRM \cite{PRM}), where $\boldsymbol{p}_{s}^{w}$ and $\boldsymbol{p}_{g}^{w}$ are start and goal poses, respectively. Then we use Algorithm~\ref{algo.IGPG} to generate an intermediate goal pose array $G^{w}$, where each pose can lie within the FOV of the camera. $r$ is the distance measurement range of the camera, and we set $r=10m$ in this paper. In Algorithm~\ref{algo.IGPG}, $\textbf{Visible}\left(\boldsymbol{m}, \boldsymbol{n}\right)$ determines whether pose $\boldsymbol{m}$ is within the FOV of the camera when the robot is located at $\boldsymbol{n}$. $\textbf{Dist}\left(\boldsymbol{m}, \boldsymbol{n}\right)$ is the euclidean distance between $\boldsymbol{m}$ and $\boldsymbol{n}$. With $G^{w}$ transformed from the world frame to the projected body frame, our mapless S2P2 can be used as a local planner given as input RGB-D images and the intermediate goal poses in $G^{pb}$ incrementally until the robot reaches the input goal. Fig.~\ref{fig.physical} shows two example trajectories from real-world experiments by using an existing navigation system with S2P2 integrated, where orange points represent the intermediate goals generated by Algorithm~\ref{algo.IGPG}.

\section{Experimental Results and Discussions}

\subsection{Datasets and Implementation Details}
We use the RGB-D dataset from \cite{wang2019self}, which covers 30 common scenes where robotic wheelchairs usually work. The input images are downsampled to $224 \times 320$ for efficiency. For our PPG, we employ four different traditional path planning algorithms: two complete algorithms, A* \cite{Astar} and JPS \cite{JPS}; and two sampling-based algorithms, PRM \cite{PRM} and RRT* \cite{RRTstar}. Each PPG generates a total number of 26429 self-supervised planned path labels given as input RGB-D images and randomly sampled goal poses. The 26429 planned path labels are split into training, validation and test sets, which contain 15859, 5285 and 5285 samples, respectively. Moreover, we adopt $\lambda_{ER} = 0.10$ and $\lambda_{IR} = 0.15$ in our experiments. For all networks, we use the stochastic gradient descent (SGD) optimizer and adopt an initial learning rate of $10^{-4}$.

We use two metrics, of which the most important is success rate (SR). SR is defined as the ratio of the number of successfully generated paths and total generated paths. A successfully generated path is defined as a path that reaches the input goal pose without colliding with obstacles. Inspired by \cite{tsardoulias2016review}, we also adopt turning cost (TC) to measure the smoothness of each path, as follows:
\begin{equation}
   TC=\frac{\sum_{i=1}^{25} \left|\theta_{i}^{pb}\right|}{25 \times 90^{\circ}},
\end{equation}
where $\theta_{i}^{pb}$ denotes the turning angle at $\boldsymbol{p}_{i}^{pb}$.

\begin{figure}[t]
   \centering
   \includegraphics[width=0.99\linewidth]{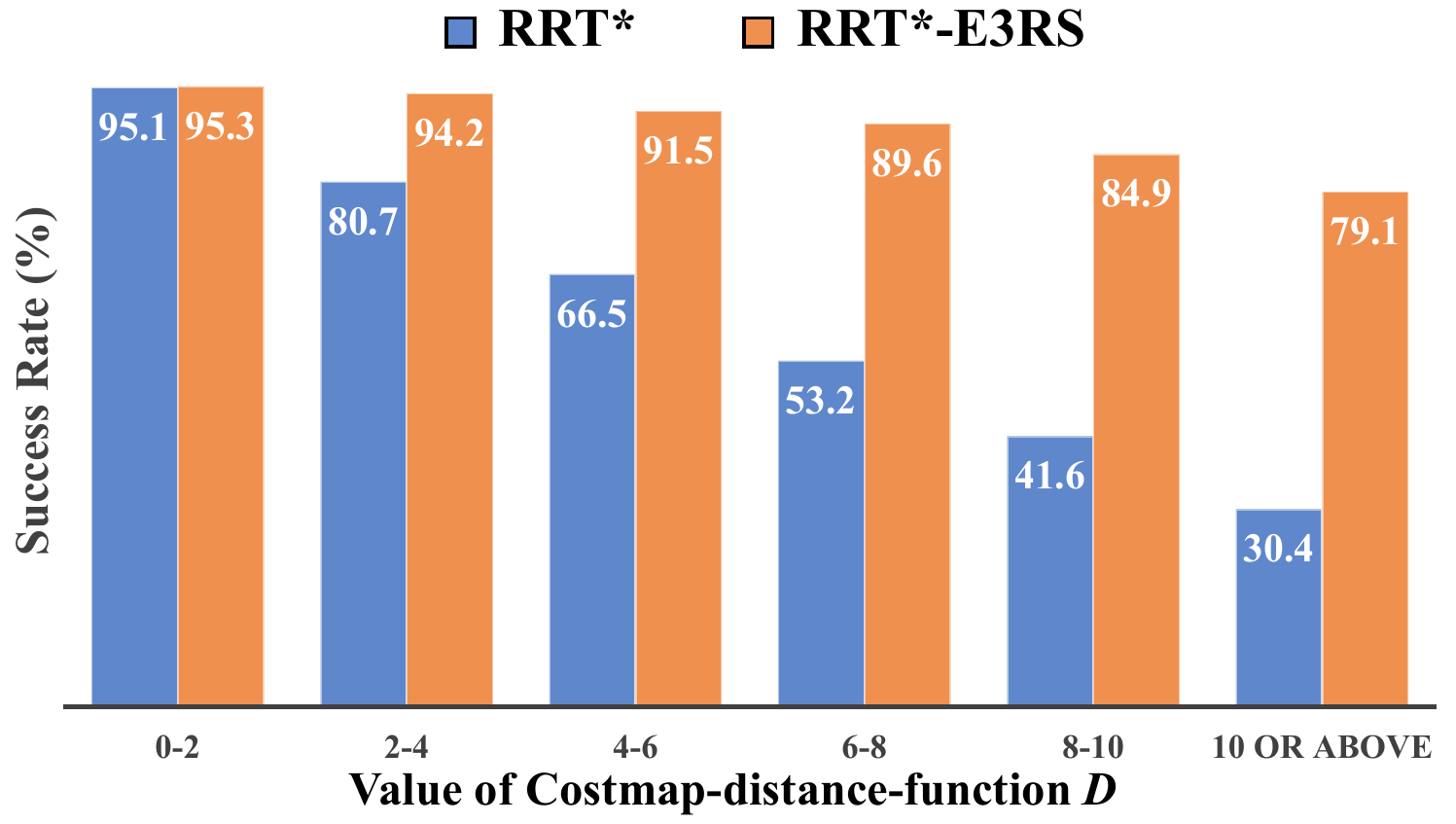}
   \caption{Comparison of the SR between RRT* and RRT*-E3RS when the quality of the perceived costmaps changes.}
   \label{fig.comparison}
\end{figure}

\subsection{Path Planning Results}

\begin{table}[t]
   \centering
   \caption{Experimental results of different variants. $\mathcal{L}_{R}$ denotes $\mathcal{L}_{ER}$ and $\mathcal{L}_{IR}$ for our E3RS and I3RS, respectively. $\uparrow$ and $\downarrow$ mean higher and lower numbers are better, respectively. The best results for our E3RS and I3RS are both bolded.}
   \begin{tabular}{C{0.4cm}L{3.7cm}C{0.6cm}C{0.9cm}C{0.9cm}}
   \toprule
   No. & Architecture & $\mathcal{L}_R$ & SR $\uparrow$ & TC $\downarrow$ \\ \midrule
   (a) & A*-PPG & -- & 70.7$\%$ & 0.132 \\ \midrule
   (b) & A*-E3RS & -- & 68.6$\%$ & 0.139 \\
   (c) & A*-E3RS (\textbf{Adopted}) & \cmark & \textbf{89.2$\%$} & \textbf{0.115} \\ \midrule
   (d) & A*-I3RS & -- & 71.8$\%$ & 0.130 \\
   (e) & A*-I3RS (\textbf{Adopted}) & \cmark & \textbf{90.2$\%$} & \textbf{0.091} \\ \midrule
   (f) & A*-I3RS (Only RGB Images) & \cmark & 83.1$\%$ & 0.124 \\
   (g) & A*-I3RS (Only Depth Images) & \cmark & 81.4$\%$ & 0.119 \\ \midrule
   (h) & A*-I3RS (One-to-Many LSTM) & \cmark & 77.2$\%$ & 0.128 \\
   (i) & A*-I3RS (FCN) & \cmark & 76.3$\%$ & 0.155 \\
   \bottomrule
   \end{tabular}
   \label{tab.ablation}
\end{table}

Jasour \textit{et al.} \cite{jasour2019risk} proposed a novel risk-contour map (RCM), a state-of-the-art approach for path planning under obstacle uncertainties. We test this approach with two different risk bounds in our test set. Table~\ref{tab.result} presents the evaluation results. It is evident that our E3RS and I3RS present significant improvements compared with the other approaches. Note that although RCM with a low risk bound performs well in SR, it presents a much worse performance in TC than E3RS and I3RS because it can generate detours easily to meet the low-risk-bound requirement. To analyze why our E3RS and I3RS can perform better than the PPG, we present some experimental results in (a)--(e) of Table~\ref{tab.ablation}. We can clearly observe that our best-fit regression plane loss $\mathcal{L}_R$ can effectively reduce the adverse impact of off-plane points and further improve the performance of planned paths.

We also choose RRT* and our RRT*-E3RS for further analysis. We use the costmap-distance-function $D$ in \cite{birk1996learning} to measure the difference between the perceived costmap constructed by the predicted semantic segmentation image and the ground-truth costmap constructed by the ground-truth semantic segmentation label. The larger the value of $D$, the worse the quality of the perceived costmap. Then, we divide the test set into six categories based on the quality of the perceived costmaps, and test the SR of RRT* and RRT*-E3RS on each category. Fig.~\ref{fig.comparison} shows that the SR of RRT* decreases much more rapidly than our RRT*-E3RS when the quality of the perceived costmap drops, which is also confirmed by the qualitative results shown in Fig.~\ref{fig.result}. We can see that when the perception results are not very accurate, our E3RS and I3RS still present a better performance than the other approaches. The reason is that the other approaches depend on the perception results, while our S2P2 is an end-to-end approach that does not. In addition, our best-fit regression plane loss $\mathcal{L}_R$ can effectively improve the performance of planned paths.

\subsection{Ablation Study}
We perform ablation studies on our A*-I3RS, which presents the best performance in our I3RS. We first test the network structures with only one kind of input, either RGB or depth images. The results in (e)--(g) of Table~\ref{tab.ablation} demonstrates the superiority of using RGB-D images \cite{wang2021pvstereo}. We speculate that it is promising to fuse RGB images with other modalities of data, such as surface normal \cite{wang2020applying,fan2021three} and optical flow \cite{wang2020cot,wang2020atg}, for autonomous navigation. To show the effectiveness of our many-to-many LSTM model, we replace the many-to-many LSTM model with two different models, a one-to-many LSTM model and a fully connected network (FCN). We can see that our many-to-many LSTM model achieves the best results from (e), (h) and (i) of Table~\ref{tab.ablation}.

\subsection{Navigation Experiments with Our Robotic Wheelchair}
\begin{figure}[t]
   \centering
   \includegraphics[width=0.99\linewidth]{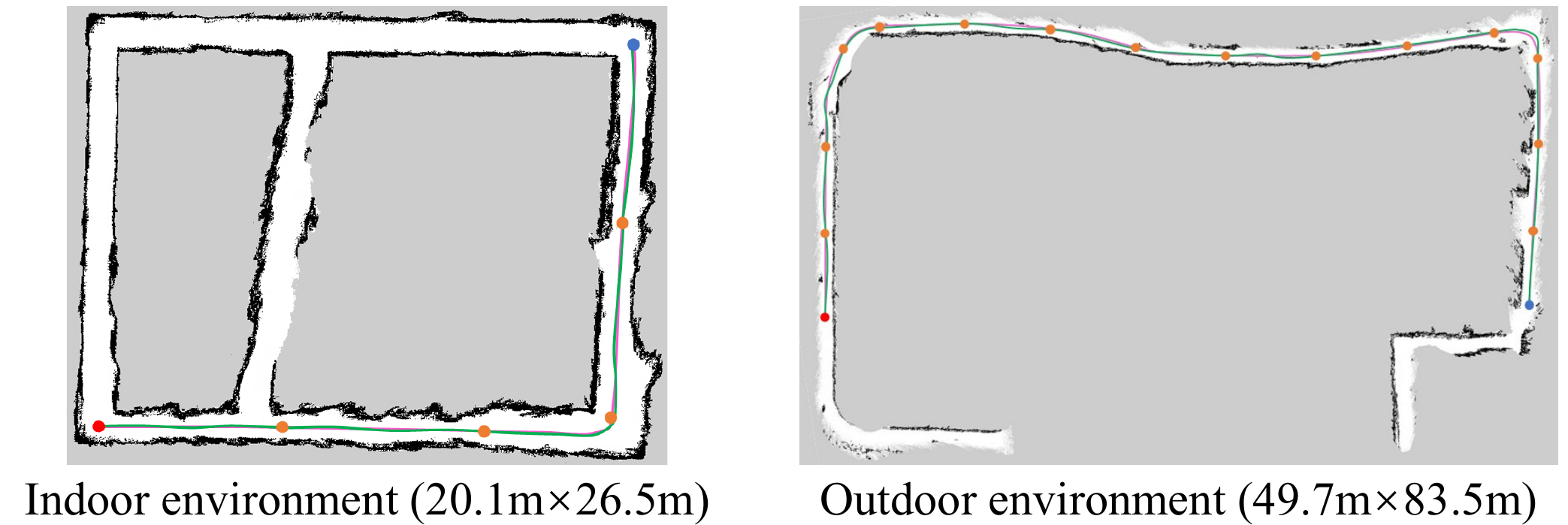}
   \caption{Two examples from real-world experiments using PRM-S2P2. Green and magenta lines denote the actual robot path and the path planned by the global planner PRM, respectively. Red, blue and orange points denote the start, the goal and the intermediate goals generated by Algorithm~\ref{algo.IGPG}, respectively.}
   \label{fig.physical}
   \vspace{-1em}
\end{figure}

To test the performance of the existing navigation system integrated with our S2P2, we use our robotic wheelchair to perform navigation tasks in one indoor environment and one outdoor environment, respectively. We choose our best approach RRT*-E3RS for our S2P2. The RGB-D SLAM system RTAB-Map \cite{labbe2013appearance} is adopted for mapping and localization. Additionally, we use PRM \cite{PRM} as the global path planner and optimal time allocation \cite{gao2018optimal} as the trajectory generator. The robotic wheelchair is commanded to track the trajectory. We call this navigation system PRM-S2P2. Fig.~\ref{fig.physical} shows two example trajectories in real-world environments using our PRM-S2P2. We can see that the robotic wheelchair can reach the goal successfully without colliding with obstacles in both the indoor and outdoor environments. As aforementioned, our S2P2 behaves as a local planner, and therefore we also compare the performance between our PRM-S2P2 and the same navigation system with other local planners integrated. The results in Table~\ref{tab.physical} demonstrate the superiority of our PRM-S2P2 over PRM-APF \cite{chiang2015path} and PRM-DWA \cite{fox1997dynamic}.

\begin{table}[t]
   \centering
   \caption{Performance comparison between different navigation systems in real-world experiments with our robotic wheelchair. The best results are bolded.}
   \begin{tabular}{L{3.3cm}C{2cm}C{2cm}}
   \toprule
   Approaches & SR (Indoor) & SR (Outdoor) \\ \midrule
   PRM-APF \cite{chiang2015path} & 60$\%$ & 50$\%$ \\
   PRM-DWA \cite{fox1997dynamic} & 65$\%$ & 55$\%$ \\
   PRM-S2P2 (\textbf{Ours}) & \textbf{95$\%$} & \textbf{90$\%$} \\
   \bottomrule
   \end{tabular}
   \label{tab.physical}
\end{table}

\section{Conclusions}
In this paper, we proposed S2P2, a self-supervised goal-directed path planning approach for robotic wheelchairs. Experimental results have demonstrated that our S2P2 outperforms traditional path planning algorithms, and increases the robustness of existing map-based navigation systems. One limitation is that the proposed approach does not explicitly model moving obstacles. Therefore, in the future, we will incorporate the moving obstacle model into our S2P2, to enable the mobile robot to present more robust and accurate navigation performance in dynamic environments.

\clearpage

\bibliographystyle{IEEEtran}
\bibliography{bibfile}

\end{document}